\documentclass[runningheads]{llncs}
\usepackage{graphicx}
\usepackage{amsmath,amssymb} 
\usepackage{color}
\usepackage{url}
\usepackage{subfigure}
\usepackage[width=122mm,left=12mm,paperwidth=146mm,height=193mm,top=12mm,paperheight=217mm]{geometry}
\usepackage{algorithm,algcompatible,amssymb,amsmath}

\algnewcommand\algorithmicto{\textbf{to}}
\algnewcommand\RETURN{\State \textbf{return} }

\algnewcommand\algorithmicinput{\textbf{Input:}}
\algnewcommand\INPUT{\item[\algorithmicinput]}

\algnewcommand\algorithmicoutput{\textbf{Output:}}
\algnewcommand\OUTPUT{\item[\algorithmicoutput]}

\algnewcommand\algorithmicinitialize{\textbf{Initialize:}}
\algnewcommand\INITIALIZE{\item[\algorithmicinitialize]}

\newcommand{\etal}{\emph{et al.}}
\newcommand{\eg}{\emph{e.g.}}
\newcommand{\ie}{\emph{i.e.}}

\usepackage[pagebackref=true,breaklinks=true,letterpaper=true,colorlinks,bookmarks=false]{hyperref}

\begin{document}
\pagestyle{headings}
\mainmatter

\title{Self-produced Guidance for Weakly-supervised Object Localization} 

\author{
Xiaolin Zhang\inst{1},
Yunchao Wei\inst{2},
Guoliang Kang\inst{1},
Yi Yang\inst{1},
Thomas Huang\inst{2}
}
\authorrunning{Xiaolin Zhang \emph{et al.}}
%

\institute{CAI, University of Technology Sydney, NSW, Australia
\email{\{Xiaolin.Zhang-3@student.,Guoliang.Kang@student.,Yi.Yang@\}uts.edu.au}\\
\and
University of Illinois Urbana-Champaign, IL, USA\\
\email{\{yunchao,t-huang1\}@illinois.edu}}

\maketitle

\begin{abstract}
Weakly supervised methods usually generate localization results based on attention maps produced by classification networks.
However, the attention maps exhibit the most discriminative parts of the object which are small and sparse.
We propose to generate Self-produced Guidance (SPG) masks which separate the foreground \ie, the object of interest, from the background to provide the classification networks with spatial correlation information of pixels.
A stagewise approach is proposed to incorporate high confident object regions to learn the SPG masks.
The high confident regions within attention maps are utilized to progressively learn the SPG masks.
The masks are then used as an auxiliary pixel-level supervision to facilitate the training of classification networks.
Extensive experiments on ILSVRC demonstrate that SPG is effective in producing high-quality object localizations maps.
Particularly, the proposed SPG achieves the Top-1 localization error rate of 43.83\% on the ILSVRC validation set, which is a new state-of-the-art error rate.
	
\end{abstract} 

%

\section{Introduction}
Weakly Supervised Learning (WSL) has been successfully applied on many tasks, such as object localization~\cite{zhou2015cnnlocalization,singh2017hide,dong2017dual,dong2017few,jie2017deep,wei2018ts2c,liang2015towards}, relation detection~\cite{zhang2017ppr} and semantic segmentation~\cite{wei2017object,wei2015stc,wei2018revisiting,wei2016learning,xiao2017transferable}.
WSL attracts extensive attention from researchers and practitioners
because it is less dependent on massive pixel-level annotations.
In this paper, we focus on Weakly Supervised Object Localization (WSOL) problem.

Existing WSOL methods locate target object regions using convolutional classification networks.
Classification networks recognize various kinds of objects by identifying discriminative regions of an objects.
Fully convolutional networks~\cite{2015-long} without using fully connected layers can preserve the relative positions of pixels. 
Therefore, the discovered discriminative regions can indicate the exact location of the target objects.
Zhou~\etal revisited classification networks (\eg AlexNet~\cite{krizhevsky2012imagenet}, VGG~\cite{simonyan2014very} and GoogleNet~\cite{szegedy2014going,szegedy2015going}) and proposed a Class Activation Maps (CAM) approach to find the regions of interest using only image-level supervision. 
Following~\cite{lin2013network}, CAM replaced the top fully connected layers by convolutional layers to keep the object positions and can discover the spatial distribution of discriminative regions for different classes.
The key weakness of the localization maps generated by CAM is that only the most discriminative regions are highlighted, as a result we can only locate a small part of target objects.
To cope with the weakness, Wei~\etal~\cite{wei2017object} proposed to apply additional networks for enriching object-related regions, given images of which the most discriminative regions are erased according to the attention maps from a pre-trained network. 
Moreover, 
Zhang~\etal~\cite{zhang2018adversarial} proved the CAM method can be simplified to enable end-to-end training. 
Armed with this proof, an Adversarial Complementary Learning approach was proposed in~\cite{zhang2018adversarial} by incorporating one additional classifier for mining complementary object regions, which can finally produce accurate object localization maps.
However, all these methods ignore to explore the correlations among pixels.

\begin{figure*}[t]
  \centering
  \includegraphics[width=1\textwidth]{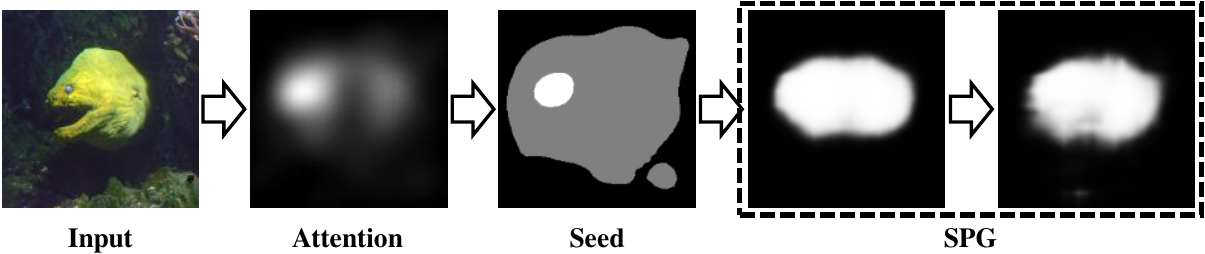}
  \caption{Learning process of Self-produced guidance. Given an input image,
  we first generate corresponding attention map with a classification network.
  Then the attention map is roughly split,
  following the rule that the region with high confidence should be the object,
  whereas that with low confidence should be background.
  The regions with medium confidence remain undefined.
  All these three regions constitute the seed.
  Self-produced guidance is defined as the multi-stage pixel-level object mask supervised by the seed.
  }
  \label{fig-0}
\end{figure*}
We observe that images can be roughly divided into foreground and background regions. 
The foreground pixels usually constitute the object(s) of interests. 
We found that attention maps inferred from classification networks~\cite{zhu2014learning,wei2017object,zhang2018adversarial} can effectively provide the probabilities of each pixel to be foreground or background. 
Although pixels of high foreground/background probabilities may not cover the entire target object/background, they still provide the important cues for getting some common patterns of target objects. 
Based on this, we can simply leverage those reliable foreground/background seeds as supervision to encourage the network to sense the distributions of foreground objects and background regions.
Since pixels with correlations (\eg within a same object or background) often share similar appearance, more reliable foreground/background pixels can be easily discovered by learning from the discovered seeds. 
With more reliable guided pixels for supervision, the entire foreground objects can be gradually distinguished from background, which will finally benefit the weakly object localization.



Inspired by the above motivation, in this paper, we propose a Self-produced Guidance (SPG) approach for learning better attention maps and getting precise positions of objects.
We leverage attention maps to produce the guidance masks of foreground and background regions in a stagewise manner.
The foreground/background seeds of each stage can be generated following a simple rule:
1) the regions with highly confident scores are considered as foreground; 
2) the regions with very low scores are background seeds; 
3) the regions with medium confidence remain undefined.
The undefined regions are meant to be figured out using intermediate features.
We adopt a top-down mechanism of using upper layer's output as the supervision of the lower layers to learn better object localizations.
The upper layers maintain more abstract semantic information, whereas the lower layers have more specific pixel-related information.
We leave the ambiguous area undefined before more regions can be defined as foreground/background using upper layer features.
The more regions be defined, the stronger ability to define harder regions.
After getting the guidance masks of foreground and background, we use them as auxiliary supervisions. 
These supervisions are expected to enable the classification network to learn pixel correlations.
Consequently, attention maps can clearly indicate class-specific object regions.
Figure~\ref{fig-0} illustrates the learning process of self-produced guidance.
Given an input image, we firstly generate corresponding attention maps through a classification network according to the convenient method in \cite{zhang2018adversarial}.
Then the attention map is roughly split into foreground/background seeds and ignored regions.
The self-produced guidance are learned from these seeds with the input of intermediate features in a stagewise manner.
Finally, the SPG masks of multiple layers are fused for a more precise and integrate indication of target objects.

To sum up, our main contributions are:
\begin{itemize}
\item We propose a stagewise approach to learn high-quality Self-produced Guidance masks which exhibit the foreground and background of a given image.
\item We present a weakly object localization method by incorporating self-produced supervision, which can inspire the classification network discover pixel correlations to improve the localization performance.
\item The proposed method achieves the new state-of-the-art with the error rate of Top-1 43.83\% on ILSVRC dataset with only image-level supervision.
\end{itemize}

We discuss the proposed SPG approach in detail in Section~\ref{sec-spg}.
In Section~\ref{exper}, we empirically evaluate the proposed method on the ILSVRC2016 dataset, showing that the superiority of SPG in object localization task with only image-level supervision.
We also discuss the further insights of the proposed SPG algorithms through additional experiments.

\section{Related Work}
Convolutional neural network has been widely used in object detection and localization tasks~\cite{simonyan2014very,jiang2013salient,luo2016accurate,he2017delving,zheng2018task,cheng2018revisiting}.
One of the earliest deep networks to detect objects in a one-stage manner is OverFeat~\cite{sermanet2013overfeat}, which employs a multiscale and sliding window approach to predict object boundaries. These boundaries are then applied for accumulating bounding boxes. 
SSD~\cite{liu2016ssd} and YOLO~\cite{redmon2016you} used a similar one-stage method, and these detectors are specifically designed for speeding up the detection process.  
Faster-RCNN designed by Ren~\etal~\cite{ren2015faster} has achieved great success in the object detection task.
It generates region proposals and predicts highly reliable object locations in an unified network in real time.
Lin~\etal~\cite{lin2017feature} presented that the performance of Faster-RCNN can be significantly improved by constructing feature pyramids with marginal extra cost.

Although these approaches are considerably successful in detecting object of interest in images, the vast number of annotations are unaffordable for training such networks with limited budget.
Weakly supervised methods alleviate this problem by using much cheaper annotations like image-level labels.
Jie~\etal~\cite{jie2017deep} proposed a self-taught learning framework by firstly selecting some high-response proposals, and then finetuning the network on the selected regions to progressively improve its detection capacity.
This method highly rely on region proposals pre-processed by algorithms like Selective Search~\cite{uijlings2013selective}. 
The general-purpose proposal algorithms may not robust to produce accurate bounding boxes. Dong~\etal~\cite{dong2017dual} adopted two separate networks to jointly refine the region proposals and select positive regions.
High-quality attention maps are also critical for object detection and segmentation~\cite{luo2018Macro}.
Diba~\etal~\cite{diba2017weakly} proposed the attention maps can be leveraged to produce region proposals.
With the assistance of these proposals, more detailed information can be easily detected.

However, these methods introduces extra computational as a result of using pre-processed region proposals and multiple networks.
Zhou~\etal~\cite{zhou2015cnnlocalization} discovered that the localization maps for each class can be produced by aggregating top-level feature maps using a class-specific fully connected layer.
Zhang~\etal~\cite{zhang2016top} introduced a different backpropagation scheme to produce contrastive response maps by passing along top-down signals downwards.
However, this method supervised by solely using image labels tends to only discover a small part of the target objects.
Wei~\etal~\cite{wei2017object} applied a similar but more efficient approach to hide discriminative regions under the guidance of a pre-trained network, and then the processed images are trained for discovering more regions of interest.
These methods increases the amount of images, thus they need much more precious computational and time resources to train the networks.
Zhang~\etal~\cite{zhang2018adversarial} provided theoretical proof of producing class-specific attention maps during the forward pass by just selecting from the last layer feature maps, which enables the end-to-end attention learning.
Also, they proposed the ACoL approach~\cite{zhang2018adversarial} to efficiently mine the integral target object in an enhanced classification network.


\section{Self-produced Guidance}\label{sec-spg}
\begin{figure*}[t]
  \centering
  \includegraphics[width=1.0\textwidth]{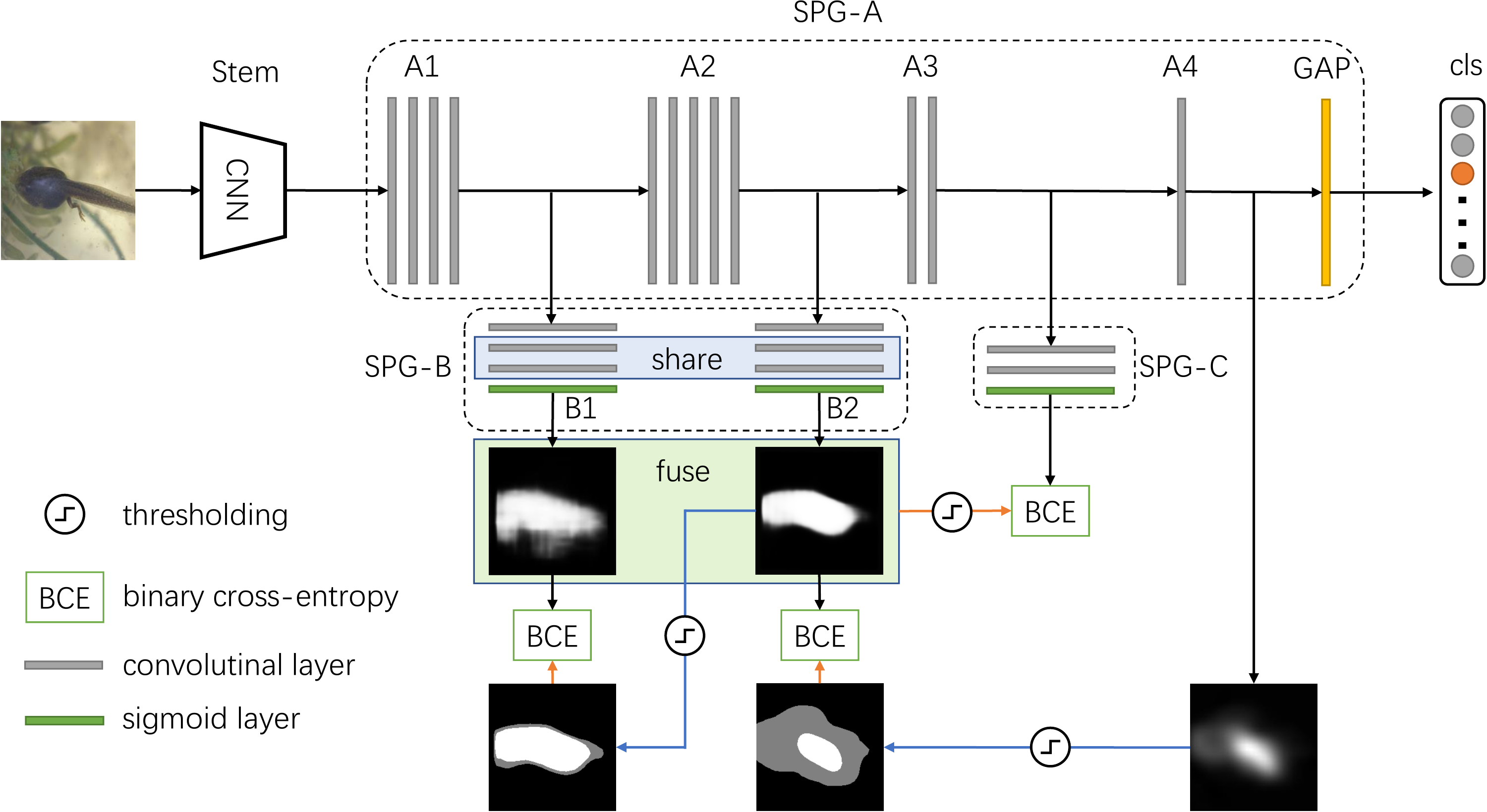}
  \caption{Overview of the proposed SPG approach. The input images are processed by Stem to extract mid-level feature maps, which are then fed into SPG-A for classification. Attention map is then inferred from the classification network. Self-produced guidance maps are gradually learned with the guide of the attention map.
SPG-C utilizes the self-produced guidance map as an auxiliary supervision to reinforce the quality of the attention map. GAP refers to global average pooling}\label{fig-1}
\end{figure*}

\subsection{Network Overview}
We denote the image set as $I=\{(I_i, y_i)\}^{N-1}_{i=0}$, where $y_i = \{0,1,...,C-1\}$ is the label of the image $I_i$, $N$ is the number of images and $C$ is the number of image classes.
Fig.~\ref{fig-1} illustrates the architecture of the SPG approach, which mainly has four different components, including Stem, SPG-A, SPG-B and SPG-C.
Different components have different structures and functionalities.
We use lowercase $f$ to denote functions and capital $F$ to denote output feature maps.
Stem is a fully convolutional network denoted as $f^{Stem}(I_i,\theta^{Stem})$, where $\theta^{Stem}$ is the parameters. The output feature maps of $f^{Stem}$ is denoted as $F^{Stem}$.
$f^{Stem}$ acts as a feature extractor, which takes the RGB images as input and produces high-level position-aware feature maps of multiple channels.
The extracted feature maps $F^{Stem}$ are then fed into the following component SPG-A.
We denote the SPG-A component as $f^{A}(F^{Stem}, \theta^{A})$, which is a network for image-level classification.
 $f^{A}(F^{Stem}, \theta^{A})$ is consisted of four convolutional blocks (\ie A1, A2, A3 and A4), a global average pooling (GAP) layer~\cite{lin2013network} and a softmax layer.
A4 has one convolutional layer with kernel size $1 \times 1$ of $C$ filters.
These filters are corresponding to the attention maps of each class, so as to generate attention maps during the forward pass~\cite{zhang2018adversarial}.
SPG-B is leveraged to learn Self-produced guidance masks by using the seeds of foreground and background generated from attention maps.
The high confident regions within attention maps are extracted to perform as supervision to learn better object regions.
SPG-B leverages the intermediate feature maps from the classification network SPG-A to predict Self-produced Guidance masks.
Particularly, the output features maps $F^{A1}$ and $F^{A2}$ of A1 and A2 are fed into the two blocks in SPG-B, respectively.
Each block of SPG-B contains three convolutional layers followed by a sigmoid layer, where the first layer is to adapt the different number of channels in feature maps $F^{A1}$ and $F^{A2}$.
The output of SPG-B are denoted as $F^{B1}$ and $F^{B2}$ for the two branches, respectively.
The component SPG-C uses the auxiliary SPG supervision to encourage the SPG-A to learn pixel-level correlations.
SPG-C contains two convolutional layers with $3 \times 3$ and $1 \times 1$ kernels, followed by a sigmoid layer.

\subsection{Self-produced Guidance Learning}
Attention maps generated from classification networks can only exhibit the most discriminative parts of target objects.
We propose to generate Self-produced Guidance (SPG) masks which separate the foreground, \ie the object of interest, from the background to provide the classification networks with spatial correlation information of pixels.
The generated SPG masks are then leveraged as auxiliary supervision to encourage the networks to learn correlations between pixels.
Thus, pixels within the same object will have the same responses in feature maps.
As the detailed information (\ie object edge and boundary) is usually very abstract in the top-level feature maps,
we employ the intermediate features to produce precise SPG masks.
Indeed, some previous works use low-level feature maps to learn object regions~\cite{xie2015holistically,hou2017deeply}.
These approaches require pixel-level ground-truth labels as supervision.
Differently, we propose to use self-produced guidance by incorporating high confident object regions within attention maps.
In detail, for any image $I_i$, we firstly extract its attention map $O$ by simply from a classification network.
We observe that the attention maps usually highlight the most discriminative regions of object.
The initial object and background seeds can be easily obtained according to the scores in the attention maps.
In particular, the regions with very low scores are considered as background, while the regions with very high scores are foreground.
The rest regions are ignored during the learning process.
We initialize the SPG learning process by these seeds.
B2 is supervised by the seed map and it can learn the patterns of foreground and background.
In this way, the pixels within the ignored regions are gradually recognized.
Then, we use the same strategy to find the foreground and background seeds in the output map of B2, which are used to train the B1 branch.
In such a stagewise way, the intermediate information of the neural network are employed to learn the Self-produced Guidance.

We formally define this process as follows. Given a input image of size $W \times H$, we denote the binarized SPG mask $M\in\{0,1,255\}^{W\times H}$,
where $M_{x,y} = 0$ if the pixel at $x_{th}$ row and $y_{th}$ column belongs to background regions, $M_{x,y} = 1$ if it belongs to object regions, and $M_{x,y} = 255$ if it is ignored.
We denote the attention map as $O$.
The produced guidance masks can be calculated by
\begin{equation}\label{eq1}
M_{x,y} = \left\{
\begin{aligned}
 0 \qquad   &if \quad O_{x,y} < \delta_l, 0<\delta_l<1 \\
 1 \qquad   &if \quad O_{x,y} > \delta_h, 0<\delta_h<1 \\
 255 \qquad &if \quad \delta_l \leq  O_{x,y}  \leq \delta_h, 0<\delta_l<\delta_h<1
\end{aligned}
\right.
\end{equation}
where $\delta_l$ and $\delta_h$ are thresholds to identify regions in localization maps as background and foreground, respectively.

We adopt an stagewise approach to gradually learn the high-quality self-produced supervision maps.
B2 is applied to learn better self-produced maps supervised by the seed map $M^{A}$.
In training, only the positions labeled as $0$ and $1$ in the self-produced maps are served as pixel-level supervision.
The pixels with values of $255$ are temporarily ignored. The ignored pixels do not contribute to the loss and their gradients do not back-propagated.
The network will learn the patterns from the already labeled pixels and then more regions will be recognized,
because the pixels belonging to background or objects usually share much correlation.
For example, the regions belong to the same object usually have the same appearance.
The output of B2 is then further applied as attention maps, and better self-produced supervision masks can be calculated using the same policy in Eq.~\eqref{eq1}.
After obtaining output maps of B1 and B2, these two maps are fused to generated our final self-produced supervision map. Particularly, we calculate the average of the two maps, then generate the self-produced guidance $M^{fuse}$ according to Eq.~\eqref{eq1}.

The generated self-produced guidance is leveraged as pixel-level supervision for the classification network SPG-A.
Thereby, the classification network will learn the correlation among pixels, and we will obtain better localization maps.
The entire network is trained in an end-to-end manner.
We adopt the cross-entropy loss function for the classification learning and self-produced guidance learning.
Algorithm \ref{alg1} illustrates the training procedure of the proposed SPG approach.

\begin{algorithm}
\small
\caption{Training algorithm for SPG}
\label{alg1}
\begin{algorithmic}[1]
\INPUT Training data $\emph{I}=\{(I_i, y_i)\}_{i=1}^N$, threshold $\delta_l$ and $\delta_h$
\WHILE {training is not convergent}
\STATE Update feature maps $F^{A4} \leftarrow  f^{A}(f^{Stem}(I_i,\theta^{Stem}),\theta^{A})$
\STATE Extract localization map $O$ from $F^{A4}$ according to image label $y_i$
\STATE Calculate the seeds of foreground/background $M^A$ according to Eq.~\eqref{eq1}
\STATE Generate the SPG map $F^{B2} \leftarrow f^{B2}(F^{A2}, \theta^{B2})$
\STATE Calculate the next stage SPG maps $F^{B1}$
\STATE Calculate the fused maps $F^{fuse}$ by averaging $F^{B1}$ and $F^{B2}$
\STATE Calculate the fused SPG masks $M^{fuse} \leftarrow F^{fuse}$ according to Eq.~\eqref{eq1}
\STATE Update the entire network $\theta_{Stem}, \theta^A, \theta^B$ and $\theta_C$ supervised by $M$ and $y_i$
\ENDWHILE
\OUTPUT Output the localization map $O$
\end{algorithmic}
\end{algorithm}

During testing, we extract the attention maps according to the class with the highest predicted scores, and then resize the maps to the same size with the original images by bilinear interpolation.
For a fair comparison, we apply the same strategy utilized in \cite{zhou2015cnnlocalization} to produce object bounding boxes based on the generated object localization maps.
In particular, we firstly segment the foreground and background by a fixed threshold.
Then, we seek the tight bounding boxes covering the largest connected area in the foreground pixels.
The thresholds for generating bounding boxes are adjusted to the optimal values using grid search method.
For more details please refer to \cite{zhou2015cnnlocalization}.

\subsection{Implementation details}
We evaluate the proposed SPG approach by modifying the Inception-v3 network~\cite{szegedy2016rethinking}.
In particular, we remove the layers after the second Inception block,~\ie, the third Inception block, pooling and linear layer.
For a fair comparison, we build a plain version network, named SPG-plain.
We add two convolutional layers of kernel size $3 \times 3$, stride 1, pad 1 with 1024 filters and a convolutional layer of size $1 \times 1$, stride 1 with 1000 units (200 for CUB-200-2011).
Finally, a GAP layer and a softmax layer are added on the top.
We update the plain network by adding two components (SPG-B and SPG-C).
The first layers of B1 and B2 are convolutional layers of kernel size $3 \times 3$ with 288 and 768 filters, respectively.
The second layers are convolutional layers of 512 filters followed by a $1 \times 1$ convolutional output layer.
The second and third layers share parameters between B1 and B2.
The strides are 1 for all convolutional layers.
To keep the resolution of feature maps, we set the pad to 1 to the filters whose kernel size is $3 \times 3$.
SPG-C is consist of two convolutional layers of kernel size $3 \times 3$ with 512 filters and a output convolutional layer with kernel size of $1 \times 1$.
All branches in SPG-B and SPG-C connects to a output sigmoid layer.
We use the pre-trained weights on ILSVRC \cite{ILSVRC15}.
Following the baseline methods~\cite{zhou2015cnnlocalization,singh2017hide}, input images are randomly cropped to $224 \times 224$ pixels after being reshaped to the size of $256 \times 256$.
During testing, we directly resize the input images to $224 \times 224$.
For classification results, we average the class scores from the softmax layer with 10 crops (4 corners plus center, same with horizontal flip).

We implement the networks using PyTorch.
We finetune the networks with the initial learning rate of 0.001 (0.01 for the added layers) on ILSVRC,
and it is decreased by a factor of 10 after every epoch.
The batch size is 30 and the weight decay is 0.0005. The momentum of the SGD optimizer is set to 0.9.
We randomly sample some images and visualize their localization maps.
We adjust $\delta_h$ to mine object seeds.
The object seeds should include as much object pixels as possible while exclude background pixels.
Similarly, $\delta_l$ can be adjusted so that the background seeds should be as large as possible while exclude object regions.
We choose the parameters for B1 are $\delta_h=0.5$ and $\delta_l=0.05$, and the parameters for B2 are $\delta_h=0.7$ and $\delta_l=0.1$.
We train the networks on NVIDIA GeForce TITAN 1080Ti GPU with 11GB memory.
Code is available at \url{https://github.com/xiaomengyc/SPG}.

\section{Experiments}\label{exper}
\subsection{Experiment setup}
\textbf{Dataset and evaluation}
We evaluate the Top-1 and Top-5 localization accuracy of the proposed approach.
We mainly compare our approach with other baseline methods on the ILSVRC 2016 dataset, as it has more than 1.2 million images of 1,000 classes for training.
We report the accuracy on the \textit{validation} set of 50,000 images.
We also tested our algorithm on the bird dataset, CUB-200-2011~\cite{WahCUB_200_2011}.
CUB-200-2011 contains 11,788 images of 200 categories with 5,994 images for training and 5,794 for testing.
We leverage the localization metric suggested by~\cite{ILSVRC15}.
An image has the right predicted bounding box if 1) it has the right prediction of image label; 2) and its predicted bounding box has more than 50\% overlap with the ground-truth boxes.

\begin{table}[t]\setlength{\tabcolsep}{10pt}
  \centering
  \caption{Localization error on ILSVRC validation set (* indicates methods which improve the Top-5 performance only using predictions with high scores).}\label{tab3}
  \begin{tabular}{l|c|c}
    \hline
    \hline
     Methods & top-1 err. & top-5 err. \\
    \hline
     Backprop on VGGnet \cite{simonyan2013deep} & 61.12 & 51.46 \\
     Backprop on GoogLeNet \cite{simonyan2013deep} & 61.31 & 50.55 \\
     AlexNet-GAP \cite{zhou2015cnnlocalization} & 67.19 & 52.16 \\
     VGGnet-GAP \cite{zhou2015cnnlocalization} & 57.20 & 45.14 \\
     GoogLeNet-GAP \cite{zhou2015cnnlocalization} & 56.40 & 43.00 \\
     GoogLeNet-HaS-32 \cite{singh2017hide} & 54.53 & - \\
     VGGnet-ACoL \cite{zhang2018adversarial} & 54.17 & 40.57 \\
     GoogLeNet-ACoL \cite{zhang2018adversarial} & 53.28 & 42.58 \\
    \hline
     SPG-plain & 53.71 & 41.81 \\
     SPG & 51.40 & 40.00 \\
     SPG* & 51.40 & 35.05 \\
    \hline
    \hline
  \end{tabular}
\end{table}

\begin{table}\setlength{\tabcolsep}{14pt}
  \centering
  \caption{Localization error on CUB-200-2011 test set (* indicates methods which improve the Top-5 performance only using predictions with high scores).}\label{tab4}
  \begin{tabular}{l|c|c}
    \hline
    \hline
    Methods & top-1 err. & top-5 err. \\
    \hline
    GoogLeNet-GAP \cite{zhou2015cnnlocalization} & 59.00 & - \\
    ACoL \cite{zhang2018adversarial} & 54.08 & 43.49 \\
    \hline
    SPG-plain & 56.33 & 46.47 \\
    SPG & 53.36 & 42.28 \\
    SPG* & 53.36 & 40.62  \\
    \hline
    \hline
  \end{tabular}
\end{table}

\subsection{Comparison with the state-of-the-arts}
We compare the proposed SPG approach with the state-of-the-art methods on ILSVRC validation set and CUB-200-2011 test set.

\textbf{Localization:}~Table~\ref{tab3}
illustrates the localization error of various baseline algorithms on the ILSVRC \emph{val} set.
We observe that our baseline SPG-plain model achieves 53.71 and 41.81 of Top-1 and Top-5 localization error.
Based on the SPG-plain network, the SPG strategy further reduces the localization error to Top-1 51.40 and Top-5 40.00.
We illustrate the results on CUB-200-2011 in Table~\ref{tab4}, the SPG approach achieves the localization error of Top-1 53.36\%.
Both results on ILSVRC and CUB outperform the state-of-the-art approach, ACoL~\cite{zhang2018adversarial} which applied two classifier branches to discover complementary object regions.
Following the baseline methods~\cite{zhou2015cnnlocalization,zhang2018adversarial}, we boost the Top-5 localization error by repeatedly using the predicted bounding boxes with high classification scores. 
We select two bounding boxes from the top 1st and 2nd predicted classes, and one from the 3rd class. 
By this way, the Top-5 localization error (indicated by *) on ILSVRC is improved to 35.05\%, and that on CUB-200-2011 is improved to 40.62\%.
To summarize, the improvement of the plain networks mainly attribute to the structure of the Inception-v3 network, which can capture larger object regions.
The improvement of the SPG networks attribute to the use of the auxiliary supervision. 
SPG can encourage the classification network learn more pixel-level correlations, and as a result of this, the localization performance increases.

Localization performance is restricted by the classification accuracy, because the calculation of localization overlap only conducts on images which have the correct prediction of image-level labels.
In order to break this limitation, we further improve the localization performance by combining our localization results with the state-of-the-art classification results, \ie, ResNet~\cite{he2016deep} and DPN~\cite{chen2017dual},
As shown in Table \ref{tab5}, the localization performance constantly improves with the classification results getting better.
When we use the classification results from the ensemble DPN method (ensemble of DPN-92, DPN-98 and DPN-131), which has very low classification error of Top-1 15.47\% and Top-5 2.70\%,
the localization error decreases to Top-1 43.83\%  and Top-5 29.36\%.
\begin{table}[t]\setlength{\tabcolsep}{6pt}
  \centering
  \caption{Localization/Classification error on ILSVRC validation set with the state-of-the-art classification results.}\label{tab5}
  \begin{tabular}{l|c|c}
    \hline
    \hline
    Methods & top-1 err. & top-5 err. \\
    \hline
    GoogLeNet-SPG-ResNet-50 & 48.79/26.22 & 38.93/8.47 \\
    GoogLeNet-SPG-ResNet-101 & 48.15/24.90 & 38.55/7.80 \\
    GoogLeNet-SPG-ResNet-152 & 47.92/24.39 & 38.53/7.59 \\
    \hline
    GoogLeNet-SPG-DPN-92  & 45.06/17.70 & 37.32/3.83 \\
    GoogLeNet-SPG-DPN-98  & 44.92/17.42 & 37.34/3.67 \\
    GoogLeNet-SPG-DPN-131  & 44.81/17.08 & 37.24/3.42 \\
    GoogLeNet-SPG-DPN-ensemble  & 43.83/15.47 & 36.78/2.70 \\
    GoogLeNet-SPG-DPN-ensemble*  & 43.83/15.47 & 29.36/2.70 \\
    \hline
    \hline
  \end{tabular}
\end{table}

\begin{figure*}
  \centering
  \begin{subfigure}[ILSVRC]{
  	\centering
  	\includegraphics[width=0.9\textwidth]{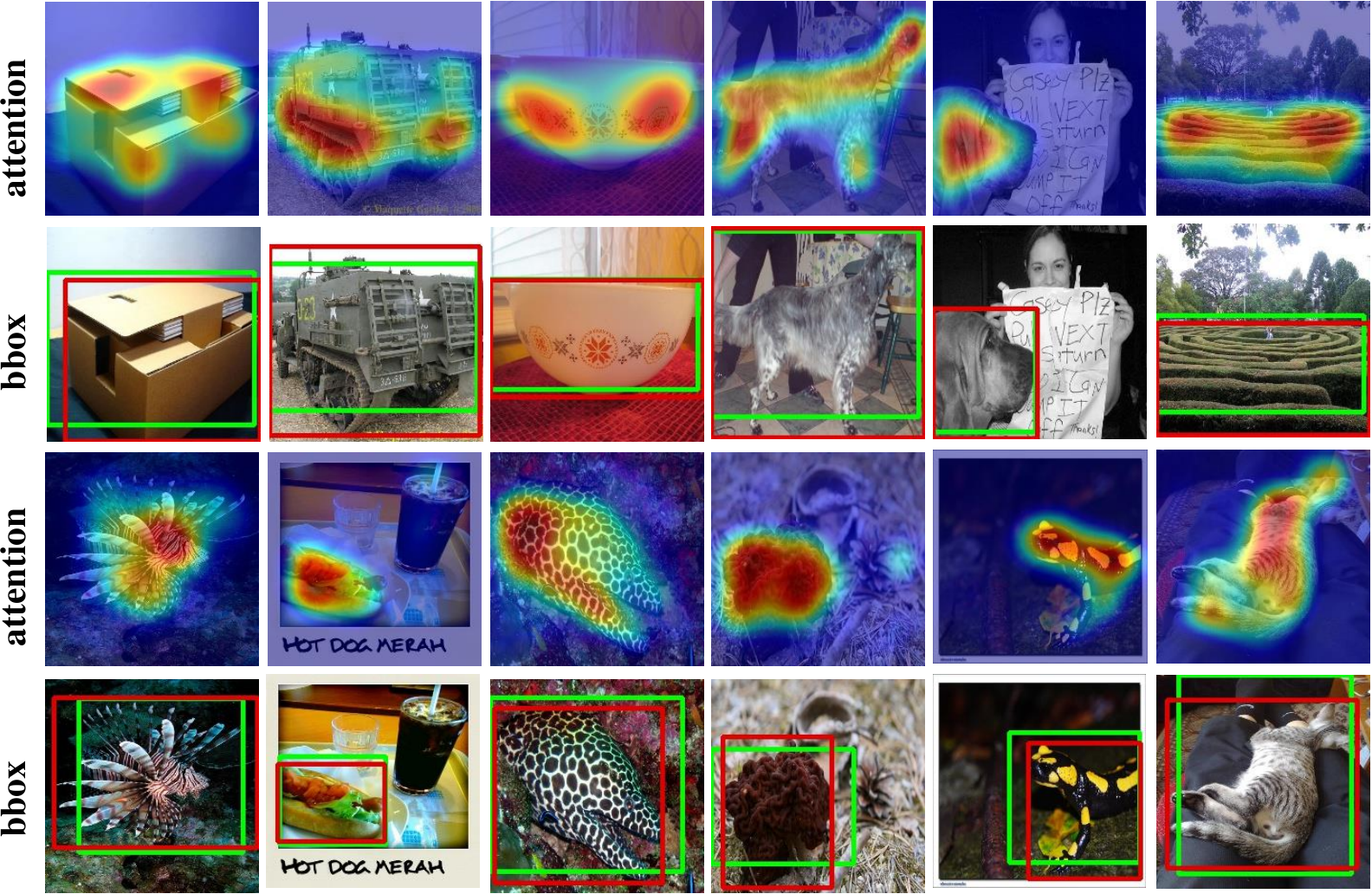}
    }
  \end{subfigure}
  \begin{subfigure}[CUB-200-2011]{
  	\centering
  	\includegraphics[width=0.9\textwidth]{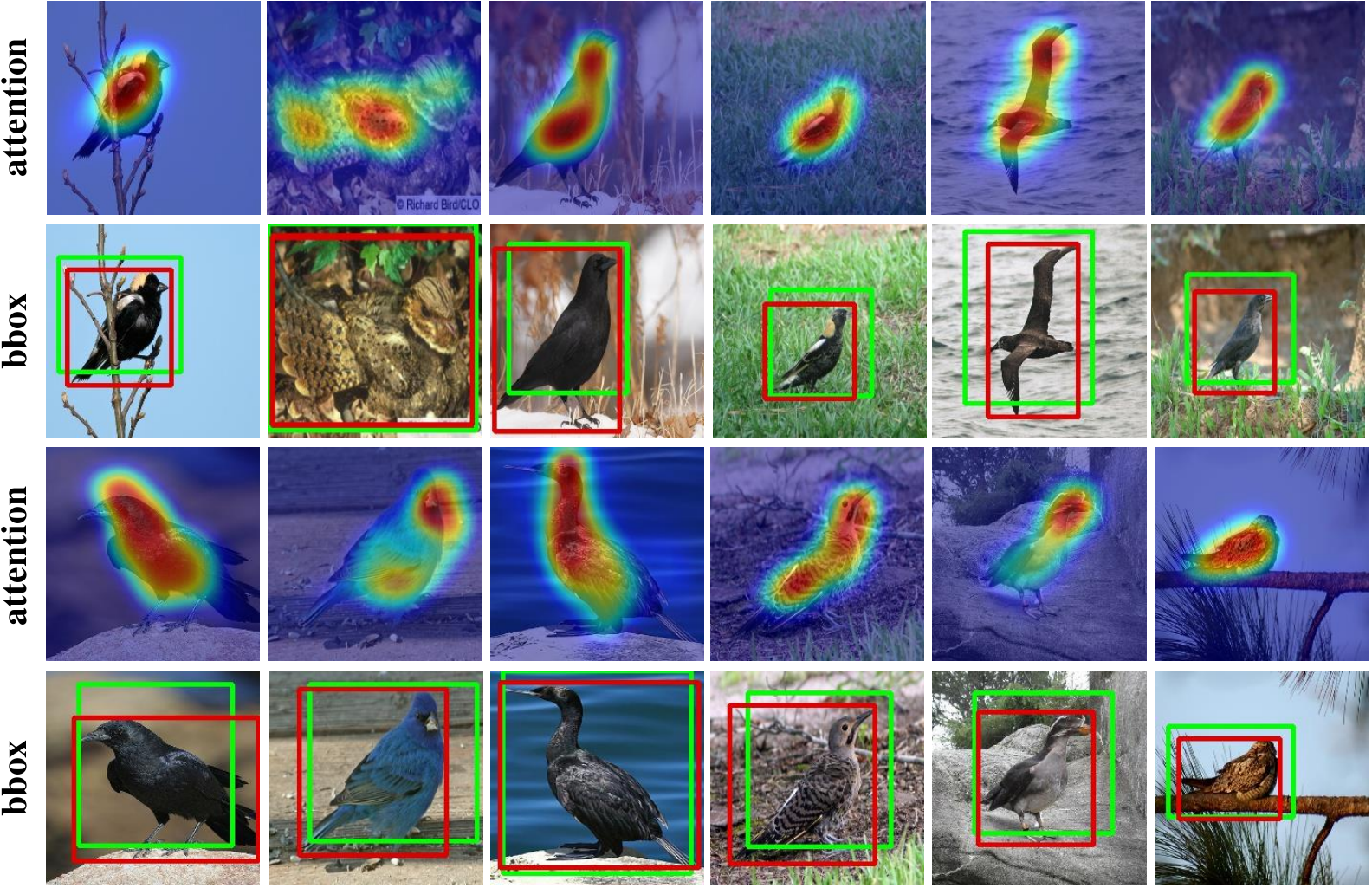}
    }
  \end{subfigure}
  \caption{Illustration of the attention maps and the predicted bounding boxes of SPG on ILSVRC and CUB-200-2011. The predicted bounding boxes are in green and the ground-truth boxes are in red. Best viewed in color.}\label{fig-box}
\end{figure*}

\begin{figure*}
  \centering
  \includegraphics[width=1.0\textwidth]{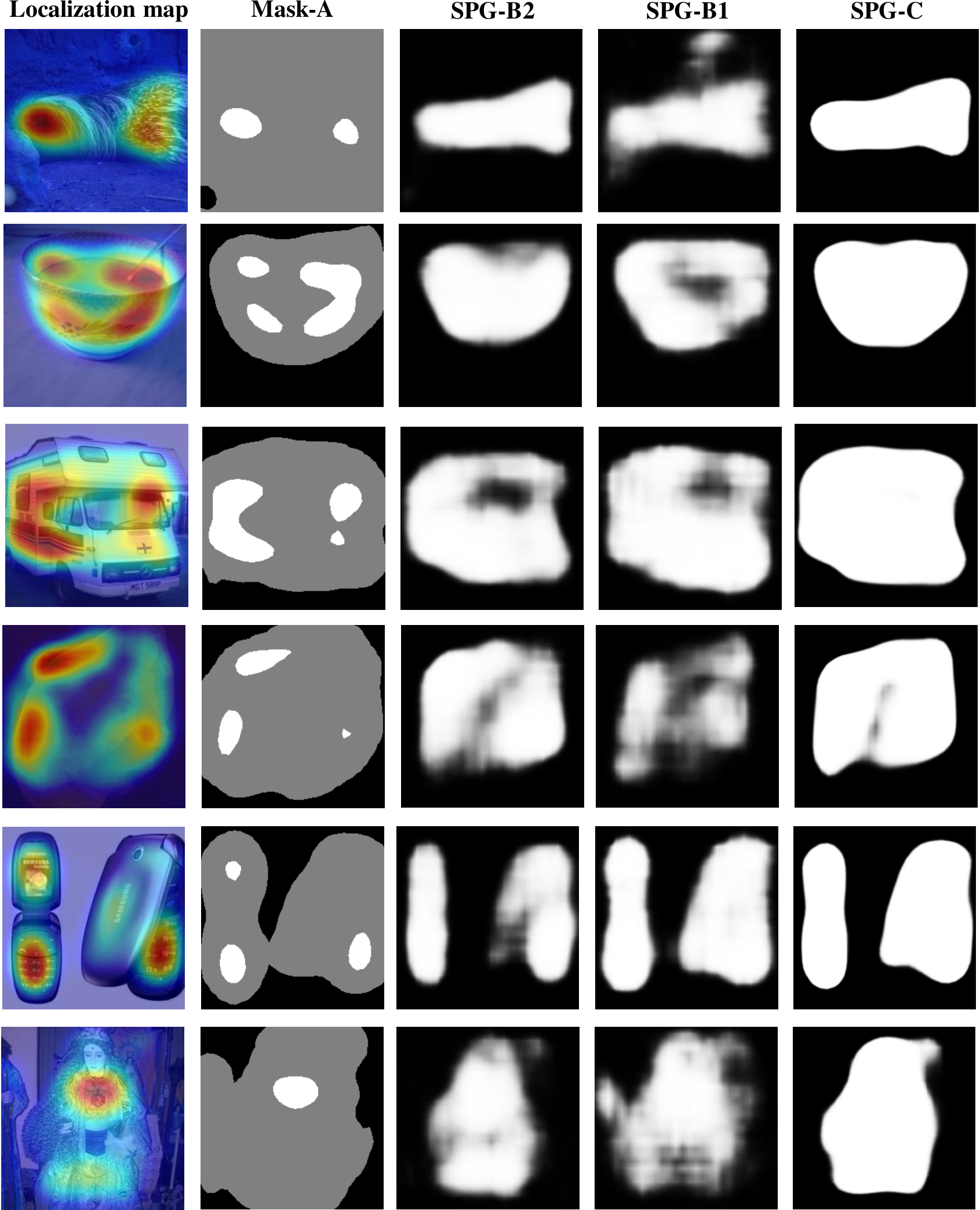}
  \caption{Output maps of the proposed SPG approach. The localization maps usually only highlight small region of the object. We extract the seeds of the self-produced guidance by segmenting the confident regions of the localization maps as foreground (white) and background (black), and ignore the left regions (grey). These seeds are applied as supervision to learn better self-produced guidance maps. Finally, the learned maps are leveraged to encourage the network to improve the quality of the localization maps.}\label{fig-2}
\end{figure*}

Figure \ref{fig-box} shows the attention maps as well as the predicted bounding boxes with the proposed SPG on ILSVRC and CUB-200-2011.
Our proposed approach can highlight nearly the entire object regions and produce precise bounding boxes.
Figure \ref{fig-2} visualizes the output of the multiple branches in generating the self-produced guidances.
The attention maps generated from the classification network are leveraged to produce the seeds of foreground and background. 
We can observe the seeds usually cover small region of the object and background pixels. 
The produced seed masks (Mask-A) are then utilized as supervision for the B2 branch.
With such supervision information, B2 can learn more confident patterns of foreground and background pixels,
and precisely predict the remaining foreground/background regions where leave undefined in Mask-A. 
B1 leverages the lower level feature maps and the supervision from B2 to learn more detailed regions.  
Finally, the self-produced guidance is obtained  by fusing the two outputs of B1 and B2.
This guidance is used as auxiliary supervision to encourage the classification network learn better attention maps.

\subsection{Ablation study}
\textbf{Limitation of the localization accuracy}

As calculation of the localization error rate is affected by network's classification performance.
We compare the localization performance using ground-truth labels to eliminate the influence caused by classification accuracy
As shown in Table \ref{tab-gtloc}, the proposed SPG outperforms the other approaches.
The Top-1 error of SPG-plain is 37.32\%, which is better than other baseline approaches.
With the assistance of the auxiliary supervision, the localization error with ground-truth labels reduces to 35.31\%.
This reveals the superiority of the attention maps generated by our method, and shows that the proposed self-produced guidance maps can successfully encourage the network learn better object regions.

\begin{table}[t]\setlength{\tabcolsep}{10pt}
  \centering
  \caption{Localization error on ILSVRC validation data with ground-truth labels.}\label{tab-gtloc}
  \begin{tabular}{l|c}
    \hline
    \hline
    Methods & GT-known loc. err. \\
    \hline
    AlexNet-GAP \cite{zhou2015cnnlocalization} & 45.01 \\
    AlexNet-HaS \cite{singh2017hide} & 41.26 \\
    AlexNet-GAP-ensemble \cite{zhou2015cnnlocalization} & 42.98 \\
    AlexNet-HaS-emsemble \cite{singh2017hide} & 39.67 \\
    GoogLeNet-GAP \cite{zhou2015cnnlocalization} & 41.34 \\
    GoogLeNet-HaS \cite{singh2017hide} & 39.43 \\
    Deconv \cite{zeiler2014visualizing} & 41.60 \\
    Feedback \cite{cao2015look} & 38.80 \\
    MWP \cite{zhang2016top} & 38.70 \\
    ACoL \cite{zhang2018adversarial} & 37.04 \\
    \hline
    SPG-plain & 37.32 \\
    SPG  & \textbf{35.31} \\
    \hline
    \hline
  \end{tabular}
\end{table}

\textbf{Effect of the cascade learning strategy}

In the proposed method, we learn the self-produced guidance maps in a two-stage way.
The branch B2 is supervised by the guidance maps generated by the localization maps from SPG-A, while the branch B1 is supervised by self-produced guidance from the output of B2.
In order to verify the effectiveness of this two-stage method, we break this structure and use the initial seed masks as supervision for the both branches.
As a result, we obtain a higher Top-1 error rate of 35.58\% when providing the ground-truth classification labels.
So, we can conclude that the two-stage structure utilized in SPG-B is useful to generate better self-produced guidance maps, and it is more effective for generating better attention maps.
Also, we find it is helpful to share the second and third layers of B1 and B2.
By removing the shared setting, the localization \textit{error rate} will increase from 35.31\% to 36.31\%.

\textbf{Effect of the auxiliary supervision}

We propose to use the self-produced guidance maps as a pixel-level auxiliary supervision to encourage the classification network to learn better localization maps using SPG-C.
Thus, we remove SPG-C to test whether SPG-C influence the classification network.
After removing SPG-C, the performance becomes worse with the Top-1 error rate of 36.06\% on ILSVRC validation set when providing ground-truth labels.
This reveals that the proposed self-produced guidance maps is effective to improve the quality of the localization maps by adding auxiliary supervision with SPG-C.
It is notable that, the localization performance with only using SPG-B is still better than the plain version.
So, the branches in SPG-B can also contribute to the improvement of localization accuracy.

\section{Conclusions}
In this paper, we proposed the Self-produced Guidance approach for locating target object regions given only image-level labels.
The proposed approach can generate high-quality self-produced guidance maps for encouraging the classification network to learn pixel-level correlations.
Thereby, the networks can detect much more object regions for localization.
Extensive experiments show the proposed method can detect more object regions and outperform the state-of-the-art localization methods.

\section*{Acknowledgement}
Xiaolin Zhang (No. 201606180026) is partially supported by the Chinese Scholarship Council.
This work is partially supported by IBM-ILLINOIS Center for Cognitive Computing Systems Research (C3SR) - a research collaboration as part of the IBM AI Horizons Network.
We acknowledge the Data to Decisions CRC (D2D CRC) and the Cooperative Research Centres Programme for funding this research.

\clearpage

\bibliographystyle{splncs}
\bibliography{egbib}
\end{document}